\documentclass[final]{cvpr}
\usepackage{times}
\usepackage{epsfig}
\usepackage{graphicx}
\usepackage{amsmath}
\usepackage{amssymb}
\usepackage[pagebackref=true,breaklinks=true,colorlinks,bookmarks=false]{hyperref}

\usepackage{amsmath,amsfonts,bm}

\def\eqref#1{equation~\ref{#1}}

\def\1{\bm{1}}

\def\rvt{{\mathbf{t}}}

\def\rvv{{\mathbf{v}}}

\def\ervt{{\textnormal{t}}}

\def\vb{{\bm{b}}}

\def\ve{{\bm{e}}}

\def\vx{{\bm{x}}}

\def\vz{{\bm{z}}}

\def\mE{{\bm{E}}}

\def\mK{{\bm{K}}}

\def\mQ{{\bm{Q}}}
\def\mR{{\bm{R}}}

\def\mV{{\bm{V}}}
\def\mW{{\bm{W}}}
\def\mX{{\bm{X}}}

\DeclareMathAlphabet{\mathsfit}{\encodingdefault}{\sfdefault}{m}{sl}
\SetMathAlphabet{\mathsfit}{bold}{\encodingdefault}{\sfdefault}{bx}{n}

\def\gC{{\mathcal{C}}}

\def\gG{{\mathcal{G}}}

\def\gL{{\mathcal{L}}}

\def\gQ{{\mathcal{Q}}}
\def\gR{{\mathcal{R}}}
\def\gS{{\mathcal{S}}}
\def\gT{{\mathcal{T}}}

\def\sR{{\mathbb{R}}}

\DeclareMathOperator*{\argmax}{arg\,max}

\usepackage{xcolor}
\usepackage{url}
\usepackage{multicol}
\usepackage{graphicx}
\definecolor{MyDarkBlue}{rgb}{0,0.08,1}
\definecolor{MyDarkGreen}{rgb}{0.02,0.6,0.02}
\definecolor{MyDarkRed}{rgb}{0.8,0.02,0.02}
\definecolor{MyDarkOrange}{rgb}{0.40,0.2,0.02}
\definecolor{MyPurple}{RGB}{111,0,255}
\definecolor{MyRed}{rgb}{1.0,0.0,0.0}
\definecolor{MyGold}{rgb}{0.75,0.6,0.12}
\definecolor{MyDarkgray}{rgb}{0.66, 0.66, 0.66}

\def\cvprPaperID{10514} 
\def\confYear{CVPR 2021}

\title{Neuro-Symbolic Representations for Video Captioning: \\ A Case for Leveraging Inductive Biases for Vision and Language}
\vspace{-0.5cm}

\author{Hassan Akbari\textsuperscript{1}\thanks{This work was conducted during the first author’s internship at MSR}, ~Hamid Palangi\textsuperscript{2}, ~Jianwei Yang\textsuperscript{2}, ~Sudha Rao\textsuperscript{2}, ~Asli Celikyilmaz\textsuperscript{2}, \\ ~Roland Fernandez\textsuperscript{2}, ~Paul Smolensky\textsuperscript{2}, ~Jianfeng Gao\textsuperscript{2}, and Shih-Fu Chang\textsuperscript{1}
\and
\normalsize \textsuperscript{1}Columbia University\\
\texttt{\tt\small\{ha2436, sc250\}@columbia.edu}
\and
\normalsize \textsuperscript{2}Microsoft Research \\\vspace{-0.5cm}
\texttt{\tt\small\{hpalangi, jianwei.yang, sudha.rao, aslicel, rfernand, psmo, jfgao\}@microsoft.com}
}

\begin{document}

\maketitle

\begin{abstract}
Neuro-symbolic representations have proved effective in learning 
structure information in vision and language. 
In this paper, we propose a new model architecture for learning multi-modal neuro-symbolic representations for video captioning. 
Our approach uses a dictionary learning-based method of learning relations between videos and their paired text descriptions.
We refer to these relations as relative roles and leverage them to make each token role-aware using attention. 
This results in a more structured and interpretable architecture that incorporates modality-specific inductive biases for the captioning task. 
Intuitively, the model is able to learn spatial, temporal, and cross-modal relations in a given pair of video and text. The disentanglement achieved by our proposal gives the model more capacity to capture multi-modal structures which result in captions with higher quality for videos. 
Our experiments on two established video captioning datasets verifies the effectiveness of the proposed approach based on automatic metrics. We further conduct a human evaluation to measure the grounding and relevance of the generated captions and observe consistent improvement for the proposed model. The codes and trained models can be found at \href{https://github.com/hassanhub/R3Transformer}{https://github.com/hassanhub/R3Transformer}. 
\end{abstract}

\vspace{-0.5cm}
\section{Introduction}
\label{intro}
\begin{figure}[!ht]
\begin{center}
    \includegraphics[width=\linewidth]{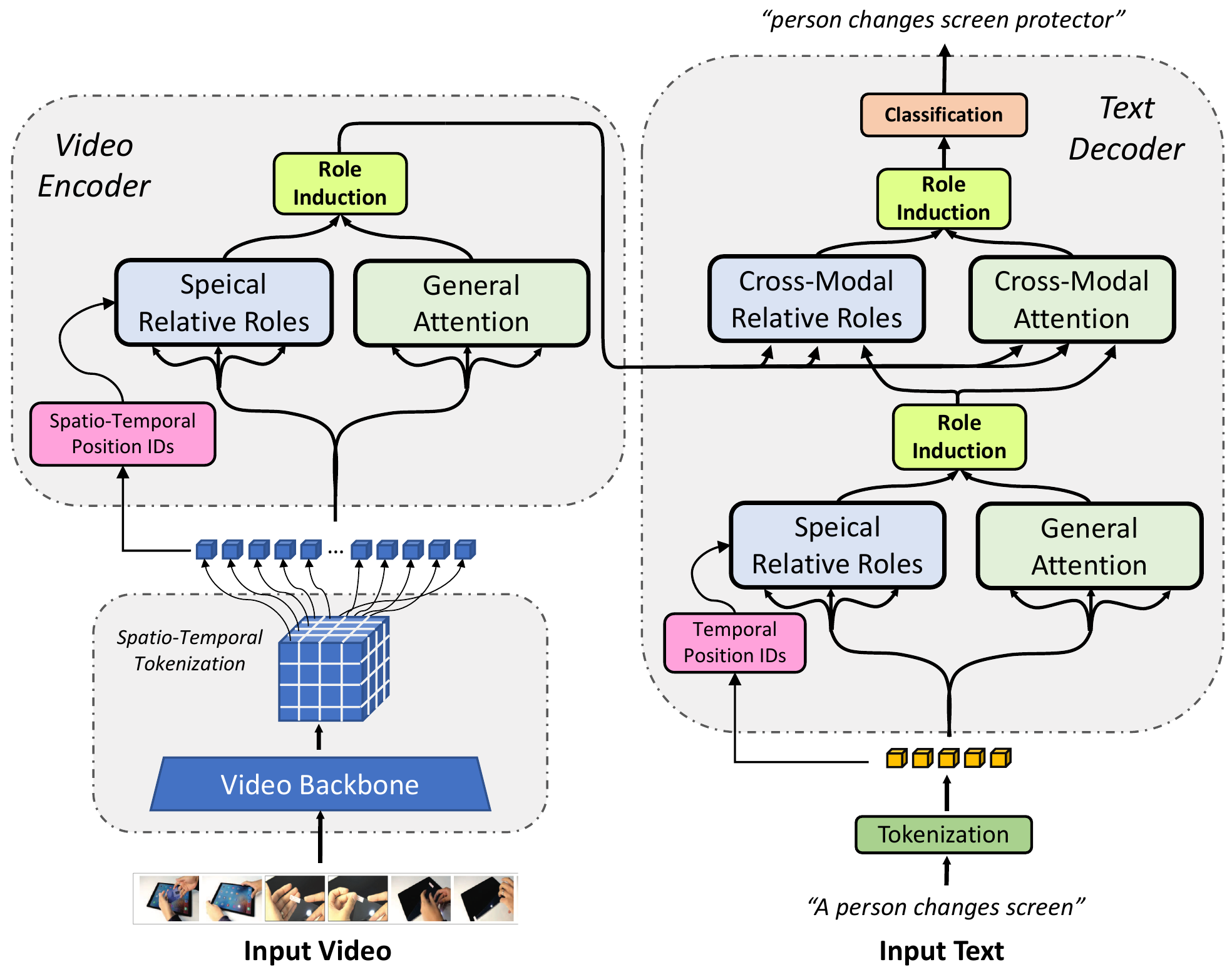}
\end{center}
\caption{High-level illustration of the proposed architecture. The key idea in this architecture is adding a module that inducts a bias toward relative roles of each token in each modality. This, beside the conventional multi-head attention (named general attention in this figure) forms a strong model that leverages such biases in multimodal data.}
\label{fig:hero}
\end{figure}
Generating natural language sentences grounded in a given context, experience or modality (e.g., grounded in vision) is of great importance and presents unique challenges \cite{bengio1,jg1,ground1,ground2}. From the initial works on image captioning (generating sentences grounded in a given scene) \cite{imcap1,imcap2,imcap3} to more modern works that leverage object detectors and neural scene graphs \cite{imcap5,hp1,imcap4} and large scale transformer based pretraining \cite{lxmert,uniter,oscar,vlp}, the common goal is to learn a unified semantic representation that captures the information from both modalities. Learning such representations is even more challenging for video captioning given the complicated nature of videos and changing dynamics among entities and their interactions over time \cite{videobert,densecap,videosurvey}. 
Significant gains reported in the literature for image or video captioning systems usually come from a combination of the following components: (i) a better vision backbone, for example the evolution from Scale-Invariant Feature Transform (SIFT) \cite{sift} to pretrained Convolutional Neural Networks (CNNs) on ImageNet, to object detectors and neural scene graphs pretrained on Stanford's Visual Genome \cite{vg} (ii) a better language backbone, for example the evolution from bag of words, to Long Short-Term Memory (LSTM) to Bidirectional Encoder Representations from Transformers (BERT) and Generative Pre-trained Transformers (GPT-(1-2-3)) \cite{bert,gpt1,gpt2,gpt3} (iii) an effective alignment mechanism, for example from an attention mechanism originally proposed for machine translation \cite{MTbengio} and leveraged for image captioning \cite{imcap2}, to self attention used in Transformers on the top text tokens (e.g., words or subwords) and visual tokens (e.g., object features). 

In this work we study the effect of a specific inductive bias for video captioning. The intuition comes from the fact that for every given dataset there are a set of \textit{objects} and \textit{actions} that carry the most important information required to complete the task. For example, in the cooking domain important objects are usually the ingredients and the actions are predicates like `frying` or `boiling`. In other domains however the set of salient \textit{objects} and \textit{actions} can be quite different and pretrained vision backbones might not offer enough coverage for object types or action types.
We want to leverage the most universal representation so that we do not need to pretrain a separate object detector for every video domain and at the same time to be able to \textit{disentangle} the actions from entities in the model for the given domain. We hypothesized (and later verified through experiments) that this type of \textit{disentanglement} will improve the performance and result in a more interpretable representation. 

To achieve this we propose to leverage an inductive bias that forces the model to learn a set of discrete \textit{roles} through vector quantization that represent the types of interaction among visual tokens. This is fundamentally based on and inspired from the inductive bias introduced in Tensor Product Representations (TPRs) \cite{ps1} in natural language processing where the hypothesis is that the grammar and semantics of a language should be decomposed and ideally represented separately.  

The proposed video captioning model includes the three common components of all captioning systems, a vision backbone for which we use voxels as our visual tokens, a language backbone for which we use transformers and variants of them, and an alignment mechanism for which we use both self attention and proposed relative role representations. Our contributions in this work are:
\begin{itemize}
    \item Introducing an inductive bias that forces the model to discretize (using vector quantization) the set of concepts (\textit{roles}) it leverages for performing the video captioning task. 
    \item Proposing a role-aware attention mechanism which helps to identify important \textit{roles} to focus on while generating captions
    \item Performing analysis on the learned \textit{roles} and identifying that in many cases they are attracted to particular predicates.
\end{itemize}

\section{Related Works}
\label{rltd_work}

The pioneering methods for video captioning mainly exploit templates~\cite{guadarrama2013youtube2text,kojima2002natural,rohrbach2013translating}. In~\cite{kojima2002natural}, the authors build a hierarchy of actions and translate the detected human action into natural language texts based on syntactic rules. \cite{guadarrama2013youtube2text} extends the hierarchy to a more semantic one to enable the description of arbitrary activities based on zero-shot recognition. In~\cite{rohrbach2013translating}, the authors exploited a conditional random field (CRF) to model the relationships between different components of video inputs, and formulate video captioning as a machine translation problem. 

In the light of deep learning techniques, many methods use a sequence-to-sequence architecture for video captioning~\cite{venugopalan2014translating,venugopalan2015sequence,pan2016jointly,xu2017learning}. In~\cite{venugopalan2015sequence}, the authors use a convolutional neural network (CNN) to extract the features of video frames and then feed them to a deep recurrent neural network for step-by-step caption generation. As an extension, S2VT~\cite{venugopalan2015sequence} uses LSTMs~\cite{hochreiter1997long} on the decoder side and the encoder side as well. To emphasize the different importance of video frames, Yao \textit{et al.} further proposed a temporal attention mechanism to attend to different temporal regions for each word generation~\cite{yao2015describing}. Besides the required coherence between a video and sentence, LSTM-E further adds another constraint to ensure relevance between them~\cite{pan2016jointly}. In this work, the authors also augment the video features with 3D CNNs~\cite{tran2015learning}. A followup work fuses three modalities --- frame, motion and audio --- to further strengthen the visual representations~\cite{xu2017learning}. Along this line, many works improve video captioning by either designing a better visual encoder~\cite{chen2018less,liu2020sibnet,chen2019temporal,aafaq2019spatio,chen2019motion} or language decoder~\cite{yu2016video,pei2019memory,wang2018video}. Of particular interest are \cite{hou2019joint, tan2020learning} which combine a sequence-to-sequence model with old-fashioned template methods for model grounding video caption generation.

Inspired by the success of the transformer~\cite{vaswani2017attention} and BERT~\cite{Devlin2018} in the language community recently, an increasing number of works resort to a transformer-based architecture for video captioning. In~\cite{zhou2018end}, the authors proposed an end-to-end dense video captioning method with a masked transformer. Similarly, \cite{chen2018tvt} replaces the transitional RNNs with transformers on both the encoder and decoder sides, while \cite{lei2020mart} proposes a memory-augmented recurrent transformer for generating more coherent dense video captions. All these works still train the models on a specific dataset. In \cite{sun2019learning}, the authors proposed VideoBert and demonstrated that pretraining a video-language representation on a large-scale dataset can significantly boost the video captioning performance. Similar findings are also discussed in some following works such as UniVL~\cite{luo2020univl} and ActBert~\cite{Zhu2020}.

In this work, we also use a transformer-based architecture. However, our model is built upon  TPRs~\cite{ps1,ps2,ps3,ps4,ps5,ps6,ps7,ps8,ps9} recently introduced into deep learning. TPRs have a specific structure that can disentangle grammar from semantics in language. However, the original TPRs do not explicitly encode the inter-token relationships and thus neglect the interplay between different parts of videos, \textit{e.g.}, the interactions between two persons. In this paper, we further propose a new module to explicitly model the interactions which proves to be particularly effective in the video captioning task.

\section{Methodology}
\label{neuro_sym}
This section introduces the proposed method and covers its various components as depicted in \autoref{fig:sr_layer} and \autoref{fig:r3tx_layer}. The proposed architecture consists of two modules, special relativity,  and general relativity. Each module will be described in detail and the proposed architecture for video captioning that leverages these modules will be presented afterwards. 

\subsection{Special Relativity}
\label{spec_rel}
\begin{figure}[t]
\begin{center}
    \includegraphics[width=6cm]{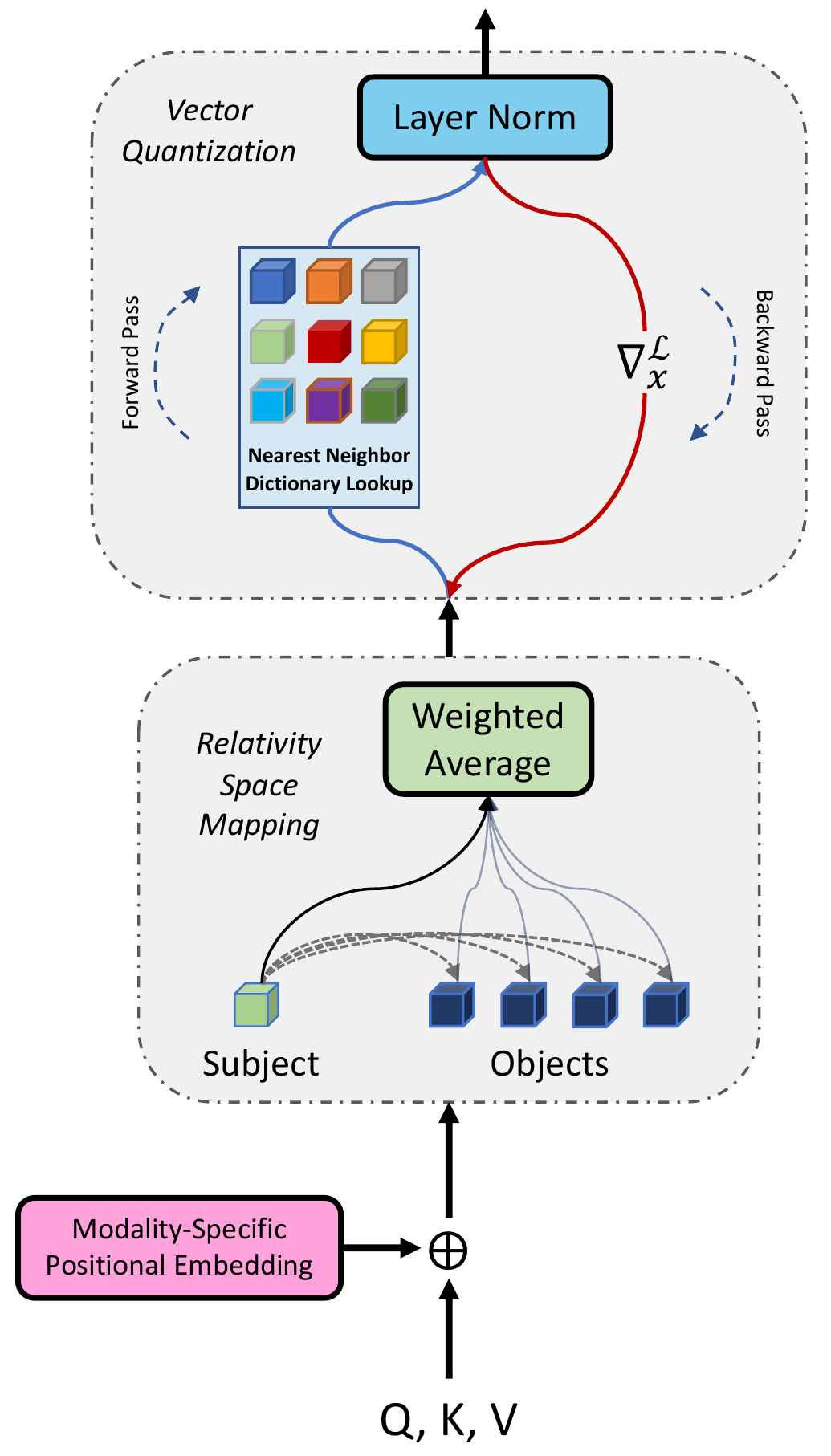}
\end{center}
\caption{An illustration of Special Relativity module. This module tries to assign a relative role (predicate) to each token (subject) with respect to its surrounding context (objects).}
\label{fig:sr_layer}
\end{figure}

The intuition behind this element is extracting relativity in a sequence of tokens (unimodal) or across tokens from a pair of sequences (cross-modal). The term ``relativity'' here refers to the very intuition of extracting the relation between a current token and its context (the rest of the tokens from either the same modality or another one). We refer to this relation as the token's role in the whole sequence and design this element in a way that specifically discovers a pre-specified number of roles and assigns them to that very token, hence we name it special relativity. In order to achieve this, we first map inputs to a space called relativity space and then learn discrete latent variables to represent each token by their role in the sequence as presented in \autoref{fig:sr_layer}.

\subsubsection{Relativity Space}
The special relativity block gets a sequence of tokens as input and adds a specific positional embedding to the tokens at each position. It then attends on the context of each token and extracts the relative representation considering these fine-grained positional embeddings. We propose to leverage spatio-temporal representations (voxels) for a given video segment as we believe spatial information plays an important role and should not be neglected through average pooling. Hence, the proposed method includes a modality-specific positional encoding and learns spatial positional embeddings as well. For the visual tokens coming from video, spatio-temporal positions are encoded as:
\begin{multline}
\label{spt_tmp_pos}
    E(t,x_0,y_0,x_1,y_1) = \\
    {\ve}_{t}^{\gT} + {\ve}_{[C_y.N].N + [C_x.N]}^{\gR\gC} + {\ve}_{[L_y.N].N + [L_x.N]}^{\gR\gS}
\end{multline}
where:
\begin{equation}
\begin{split}
    C_y = \frac{y_0+y_1}{2}, C_x = \frac{x_0+x_1}{2},\\
    L_y = |y_1-y_0|, L_x = |x_1-x_0|
\end{split}
\end{equation}
$N^2$ represents maximum number of spatial buckets (think of an $N \times N$ grid). \autoref{spt_tmp_pos} divides the frame at time $t$ into $N\times N$ regions and encodes a bounding box in the frame (represented by $[x_0,y_0,x_1,x_2]$) by assigning its center and shape to $N$ possible buckets at each dimension. Similarly, temporal locations are assigned to $T$ possible buckets. Finally, each bucket is represented by a learnable vector $e_i \in \sR^{d_{model}}$ within an embedding lookup, hence we have three learnable embedding matrices $\mE^{\gT}$, $\mE^{\gR\gC}$, and $\mE^{\gR\gS}$ for temporal, region center and region size, respectively. On the other side, in text, we only use temporal location embedding lookup and only learn the first term of Eq.\ref{spt_tmp_pos} separately and dedicated for text specifically. These representations are calculated once throughout the model and are fed into each special relativity layer.

In order to calculate relativity, the model attends to the context of each token using a conventional multi-head attention \cite{vaswani2017attention}. Inspired by the Encoder-Decoder architecture \cite{vaswani2017attention}, we propose self-relativity and cross-relativity, where in the former we find unimodal relativity representations for each token and in the latter we incorporate grounding into the realm of relativity. Assume we are feeding $\mQ_{in}$, $\mK_{in}$, and $\mV_{in}$ as inputs to the proposed layer, then the multi-head relativity is calculated for each head as follows:
\begin{equation}
\label{eq:qkv}
\begin{split}
    \mQ_{\gS}^{h} = \mQ^{h}_{in}\mW^{h}_{q},\\
    \mK_{\gS}^{h} = \mK^{h}_{in}\mW^{h}_{k},\\
    \mV_{\gS}^{h} = \mV^{h}_{in}\mW^{h}_{v}
\end{split}
\end{equation}
Per-head relativity of the given tokens is calculated as:
\begin{equation}
    \mX_{\gS}^{h} = Softmax(\frac{\mQ_{\gS}^{h}{\mK_{\gS}^{h}}^{T}}{\sqrt{d_k}})\mV_{\gS}^{h}
\end{equation}
where $d_k$ represents the dimension of the relativity and is used as a normalization factor in the Softmax function for more robust representations with respect to variance \cite{vaswani2017attention}. $\mX_{\gS}^{h}$ represents the per-head relativity for the given input tokens.

\begin{figure}[t]
\begin{center}
    \includegraphics[width=\linewidth]{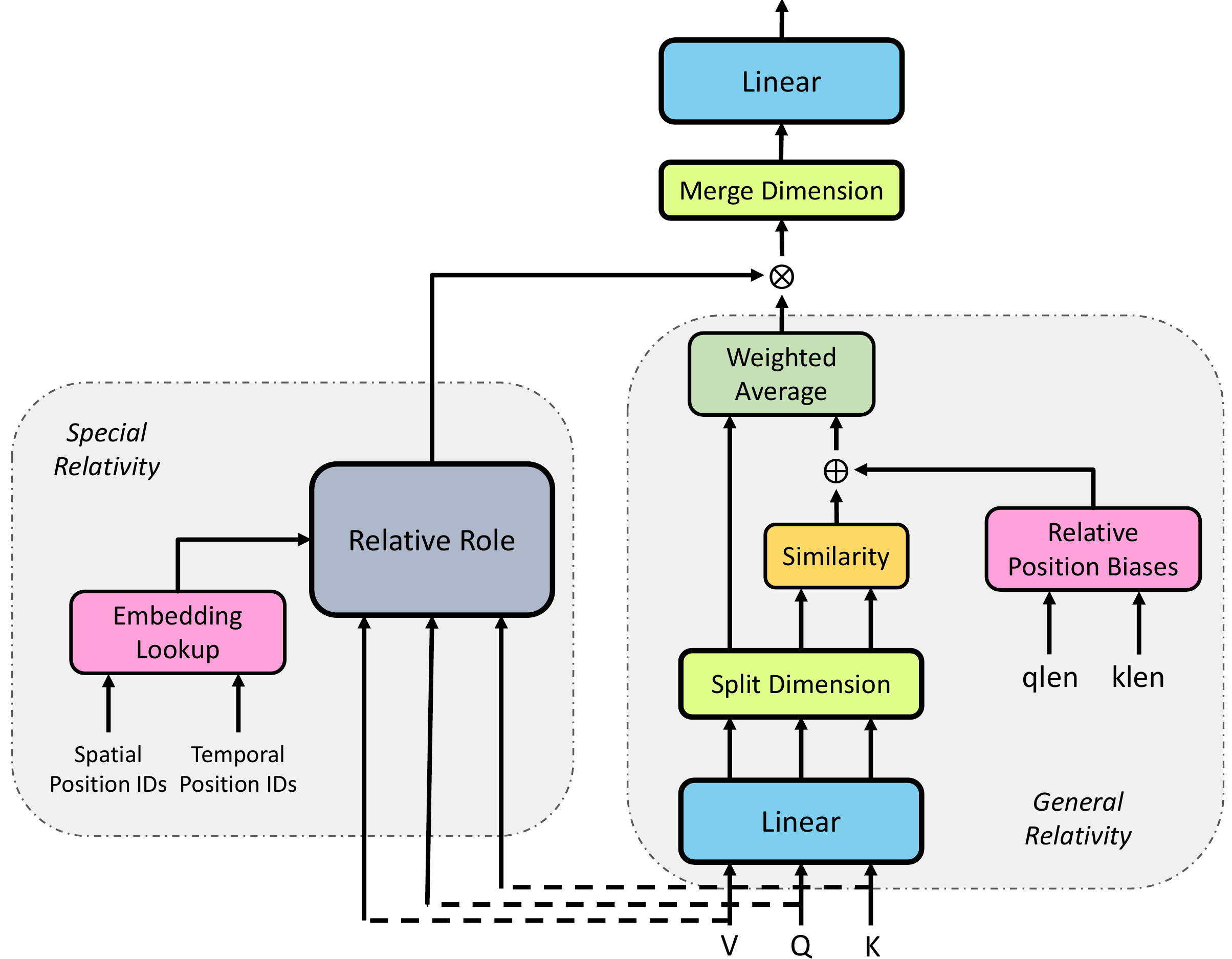}
\end{center}
\caption{Our proposed Multi-Head Attention layer for extracting Relative Role Representations (R3)}
\label{fig:r3tx_layer}
\end{figure}

\subsubsection{Discrete Latent Variables}
Once we have the representations in relativity space, the next step is learning a set of discrete latent variables, which is referred as \textit{roles} that will be learned during training. The intuition behind using discrete latent variables is to quantize the relativity space as a resemblance to a finite set of objects, subjects, and predicates in each modality. As we explain later, this concept can be extended to cross-modal relations similarly. In order to quantize the relativity space, we utilize a nearest neighbor lookup and backpropagate the gradient by straight-through estimation \cite{bengio2013estimating}. We first define a latent embedding space $\mE \in \sR^{K\times d_k}$, where $K$ is the size of the latent variable space. Equivalently, it could be seen as a $K$-way categorical representation \cite{van2017neural}.
Given a vector in the relativity space, $\vx^{h} \in \sR^{d_k}$, the discrete latent variables $\vz^{h}_{q}$ are calculated as follows:
\begin{equation}
\label{eq:vq}
    \vz^{h}_{q}(\vx^{h}) = \ve_{n} ~~; n = \argmax_{j}{\ve_{j}^{T}.\vx^{h}}
\end{equation}
This means that the posterior categorical distribution $q(z|x)$ probabilities are defined as:
\[ q(z=n|x) =
  \begin{cases}
    1       & \quad \text{for } n = \argmax_{j}{\ve_{j}^{T}.\vx^{h}}, \\
    0       & \quad \text{otherwise}
  \end{cases}
\]
More specifically, we choose a deterministic approach for calculating latent variables, however more sophisticated non-deterministic categorical reparameterizations can be utilized to better model uncertainties \cite{jang2016categorical}. We leave such approaches for future studies.

As shown in \autoref{eq:vq}, vector quantization is performed per-head and each head chooses from a shared pool of dictionary words $\ve_{j}$. Hence, the resulting concatenated vector, $\vz_{q}$, has $K^H$ number of possibilities in the quantized space. We tried per-head separate dictionary similar to \cite{baevski2019vq}, but did not observe any improvements in our final results. Moreover, we normalize vectors $\vx^{h}$ and $\ve_{j}$ before quantization to avoid variance biases during training. Using the inner product instead of euclidean distance in \autoref{eq:vq} enables us to perform dictionary lookup more efficiently on GPUs while achieving similar results.

Since \autoref{eq:vq} is not differentiable with respect to $\vx^{h}$, following \cite{van2017neural} we use straight-through estimation, where the quantized vector ($\vz^{h}_{q}$) is fed to the next layers during the forward pass while the gradients are calculated with respect to continuous vectors ($\vx^{h}$) during the backward pass. This can be achieved by re-writing \autoref{eq:vq} as:

\begin{equation}
\label{eq:sq_vq}
    \vz^{h}_{q}(\vx^{h}) = sg(\ve_{n}-\vx^{h}) + \vx^{h} ~~; n = \argmax_{j}{\ve_{j}^{T}.\vx^{h}}
\end{equation}

where $sg$ stands for StopGradient, an implementation terminology, that cuts the flow of gradients from that point during backward pass. Finally, we add \textit{commitment loss} and \textit{dictionary loss} terms to final model loss to ease the training \cite{van2017neural}:

\begin{equation}
\label{eq:vq_loss}
    \gL_{\gQ} = \underbrace{{\|sg(\vx^{h})-\ve_{n}\|_{2}^{2}}}_{\textit{dictionary loss}} + \beta \underbrace{{\|sg(\ve_{n})-\vx^{h}\|_{2}^{2}}}_{\textit{commitment loss}}
\end{equation}

Since this constraint is only applied on the \textit{chosen} dictionary word, $\ve_{n}$, and not other words in the dictionary (at each backward pass), we observed that applying dropout on the dictionary lookup similarities ($\ve_{j}^{T}.\vx^{h}$) makes model training more robust. This basically reduces the affect of initialization bias and avoids the model from propagating gradient only over a few chosen dictionary words repeatedly.

\subsection{General Relativity}
\label{gr_layer}
This module performs a multi-head attention on the input sequence and scales them by attention scores. The intuition behind using this module is contextualizing both uni-modal and cross-model representations. In contrast to Special Relativity, we put no constraint on this module, hence assuming that this module extracts the general context in input sequences and Special Relativity further garnishes those information by special relative roles. We utilize the best architecture explored in \cite{raffel2020exploring}, where $\mQ$, $\mK$, and $\mV$ are calculated using \autoref{eq:qkv} (with different learnable weights) while the per-head representations are modulated by a relative positional bias as follows:
\begin{equation}
    \mX_{\gG}^{h} = Softmax(\mQ_{\gG}^{h}{\mK_{\gG}^{h}}^{T}+\vb^{h})\mV_{\gG}^{h}
\end{equation}
where $\vb^{h}$ is a learnable bias lookup which maps a finite set of buckets to a scalar bias. These buckets are calculated based on the relative distance of the tokens in $\mQ_{\gG}^{g}$ and $\mK_{\gG}^{g}$ or the "key-query" offsets \cite{shaw2018self}. We utilize one bias lookup for encoder and two for decoder. Each of these are only calculated in the first layers and are propagated through the rest of the layers. We refer to \cite{raffel2020exploring} for more details and the benefit of these choices.

\subsection{Relative Role Representations (R3)}
\label{r3tx_layer}
\autoref{fig:r3tx_layer} shows our proposed architecture that utilizes representations from Special Relativity and General Relativity modules. We utilize a Hadamard product as an estimation of Tensor product \cite{ps3} to garnish general representations with special relative representations and achieve a TP representation as below:

\begin{equation}
\label{eq:tp-r3}
    \mR^{h} = \mX_{\gS}^{h} \odot \mX_{\gG}^{h}
\end{equation}
where $\mR^{h}$ is the per-head relative role-aware representation. The per-head representations are later merged in to one space and are followed by a linear mapping to form the final multi-head attention output, $\mR$. We call our multi-head attention module R3-MHA and build Transformer encoder and decoder layers similar to \cite{vaswani2017attention}. For the rest of the paper, we refer to our proposed architecture as R3-Transformer and call the baseline (General Relativity-only MHA) Transformer.

\section{Experiments}

\begin{table*}[t]
\begin{center}
\scalebox{0.9}{
    \begin{tabular}{|l|c|cccccccc|}
        \hline
        Method & Dataset & B@1 & B@2 & B@3 & B@4 & METEOR & ROUGE & CIDEr & SPICE  \\ \hline \hline
        \multicolumn{1}{|l|}{Bi-LSTM \cite{bilstm}} &  & - & - & - & 0.87 & 8.15 & - & - & - \\
        \multicolumn{1}{|l|}{EMT \cite{zhou2018end}} &  & - & - & - & 4.38 & \textbf{11.55} & 27.44 & 38.00 & - \\
        \multicolumn{1}{|l|}{UniVL \cite{luo2020univl}} & YouCook II & 25.73 & 13.55 & 7.35 & 4.09 & 10.01 & 27.24 & 45.89 & 13.12 \\
        \multicolumn{1}{|l|}{Transformer \cite{raffel2020exploring}} &  & 24.30 & 12.24 & 6.23 & 3.37 & 9.43 & 26.19 & 41.68 & 13.20 \\
        \multicolumn{1}{|l|}{R3-Transformer (ours)} &  & \textbf{25.74} & \textbf{13.67} & \textbf{7.61} & \textbf{4.50} & 10.19 & \textbf{28.17} & \textbf{50.22} & \textbf{14.45} \\ \hline
        \multicolumn{1}{|l|}{LSTM-YT \cite{venugopalan2015sequence}} &  & 18.22 & 7.43 & 3.24 & 1.24 & 6.56 & - & 14.86 & - \\
        \multicolumn{1}{|l|}{S2VT \cite{venugopalan2014translating}} &  & 20.35 & 8.99 & 4.60 & 2.62 & 7.85 & - & 20.97 & - \\
        \multicolumn{1}{|l|}{DCEV \cite{densecap}} & ActivityNet & 20.74 & 9.29 & 4.76 & 2.75 & 8.14 & - & 23.43 & - \\
        \multicolumn{1}{|l|}{Transformer \cite{raffel2020exploring}} &  & 21.28 & 9.86 & 5.23 & 3.06 & 8.66 & 18.94 & 20.72 & 10.87 \\
        \multicolumn{1}{|l|}{R3-Transformer (ours)} &  & \textbf{23.53} & \textbf{11.65} & \textbf{6.54} & \textbf{4.01} & \textbf{9.78} & \textbf{20.42} & \textbf{29.68} & \textbf{13.20} \\ \hline
    \end{tabular}
    }
\end{center}
    \caption{Captioning performance comparison among different methods trained on the YouCook II and ActivityNet datasets. All methods were trained from scratch and no multimodal pre-training was performed before the video captioning training task. Our results on ActivityNet are segment-based, which means no previous segments or captions were participating in generating future captions.}
    \label{tab:cap_numbers}
\end{table*}

\label{exp}
This section describes the experimental setup and how the proposed method is evaluated on two established video captioning datasets, including justifications for the choices made. Please note that each model was trained from scratch on the given dataset, there was no vision-language pretraining and each model was trained using the same captioning loss. Through our experiments the goal was to answer the following questions:
\begin{itemize}
    \item Does the proposed architecture perform better than Transformers for video captioning?
    \item Is the performance improvement observed across different domains?
    \item What type of information is stored in the learned \textit{roles}?
    \item Does the model generate better captions in the eyes of human evaluators?
\end{itemize}
\subsection{Datasets}
\subsubsection{YouCook2}
This dataset contains 2000 videos gathered from YouTube and includes 89 different recipes for cooking \cite{zhou2020unified}. The overall duration is 176 hours with an average of 5.26 minutes per video. Each video is divided into multiple segments, and each segment is annotated with a text description (caption) without any other context (e.g. previous segments). The overall number of segments is 14k, and each segment has an average duration of 30 seconds and 10 tokens. Following \cite{shi2019dense}, we split the data into 1,261 training videos and 439 test videos. This results in 9,776 training segments (video-text pairs) and 3,369 test segments.

Even though the segments have an average of 30 seconds and 10 tokens, given the histogram of these variables across all samples (shown in Appendix \ref{appndx:ablation}), we pad each segment to a maximum of 50 seconds and 50 tokens (zero pad if shorter, clip if longer).

\subsubsection{ActivityNet Captions}
The ActivityNet Captions dataset \cite{krishna2017dense} includes 100k dense natural language descriptions of 20k videos from the original ActivityNet dataset \cite{caba2015activitynet}, summing to an overall duration of 849 hours. Similar to YouCook2, this dataset has multiple segments per video. However, the descriptions for each segment considers the context from previous segments in addition to a precise timing annotation for each segment, which is the main reason that this dataset is used for dense captioning. Each segment has an average of 36 seconds and 20 tokens. We choose a maximum length of 100 seconds for video frames and 50 tokens for text descriptions (see Appendix \ref{appndx:ablation}), and use ground-truth segmentation annotations without any context from previous segments in our experiments.

\subsection{Experimental Setup}
\vspace{-0.1cm}
\subsubsection{Feature Representations}
\vspace{-0.15cm}
We normalize all video clips to 32 fps, resize the frames to $124$ on their smaller dimension, and apply a central crop of $112\times112$ on the resulting clips. In order to extract video features, we first prepare our own video backbone by fine-tuning a SlowFast-50 network \cite{feichtenhofer2019slowfast} on the Kinetics-400 dataset \cite{kay2017kinetics} with the same pre-processing approach. The video model uses $\alpha=8, \tau=8$, hence each clip fed to the network has a duration of 64 frames. Instead of using the classification features of SlowFast, we apply a temporal pooling with kernel and stride size of 4 on the temporal dimension of 3D feature maps from the Slow pathway. Similarly, we apply a temporal pooling on the outputs of the Fast pathway, but multiply the pooling kernel and stride size of $\tau$ to have a feature map with the same dimensionality as the Slow pathway. We then concatenate these features and apply a spatial average pooling with kernel size of 2 with \textit{valid} padding to form an overall 3D feature representation of size $2\times3\times3\times2304$ per clip. This feature corresponds to 9 regions across 2 seconds of video content and overall it results in 18 tokens. We studied different sampling rates and features and found these parameters to be the ones that deliver the best results on validation splits. On the text side, we use a pre-trained T5 tokenizer which tokenizes each sentence to wordpiece tokens with special tokens including "$<$/s$>$" (eos), "$<$pad$>$", and "$<$mask$>$". Beginning of sentence (bos) is coded with the "$<$pad$>$" token. We use a pre-trained embedding lookup trained for T5-base \cite{raffel2020exploring} and fine-tune it in our experiments. T5-tokenizer is based on the Huggingface implementation of T5 \cite{Wolf2019HuggingFacesTS} and performs WordPiece tokenization \cite{schuster2012japanese} for a given string.

\subsubsection{Training and Evaluation}
We implemented our models from scratch in Tensorflow 2.3, while borrowing SlowFast implementation logic from \cite{fan2020pyslowfast}, and Transformer from \cite{Wolf2019HuggingFacesTS}.

Training loss function is defined as a combination of VQ loss and causal conditional masked language modeling as follows:
\vspace{-0.1cm}
\begin{equation}
\label{eq:train_loss}
    \gL(\theta) = -E_{\ervt_i\sim\rvt}\log p_{\theta}(\ervt_i|\ervt_{<i},\hat{\rvt},\rvv) + \gL_{\gQ}
\end{equation}
where the left term is the cross-entropy loss defined based on the probability distribution from the text decoder. This can be seen as a next word prediction task given a masked text ($\hat{\rvt}$) conditioned on video ($\rvv$), hence a combination of conditional masked language modeling and language reconstruction. We tried alternative optimization of these two tasks but found no gain in final results.
We trained our model using Adam optimizer \cite{kingma2014adam} with a fixed learning rate, since warmup and decay provided no gain in our experiments. We found that each dataset requires its own training parameters for an efficient performance, hence we conducted an extensive ablation study and found best combination of parameters for each dataset. Please refer to \ref{appndx:ablation} for more details.
Evaluation is performed by first calculating the encoder states for a given video and then autoregressively generating the caption using the decoder conditioned on the video states. We use the implementation in \cite{Wolf2019HuggingFacesTS} with no repetition penalty,  length penalty, or early stopping. We cap the maximum length of the generated captions to 50 tokens (same as the training maximum length). We found a greedy decoding giving the best results (based on automatic captioning metrics).

\begin{figure*}[!ht]
\begin{center}
    \includegraphics[width=\linewidth]{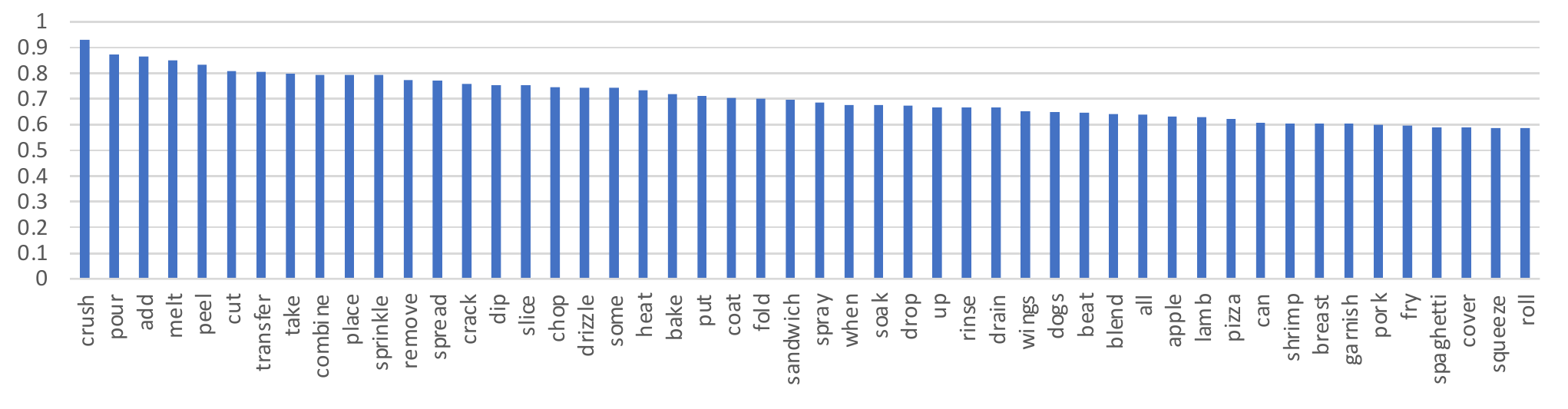}
\end{center}
\caption{Probability of selecting the same role (vertical axis) given the generated word (horizontal axis) for the YouCook II evaluation set.}
\label{fig:youcook2roles}
\end{figure*}

\begin{figure*}[!ht]
\begin{center}
    \includegraphics[width=\linewidth]{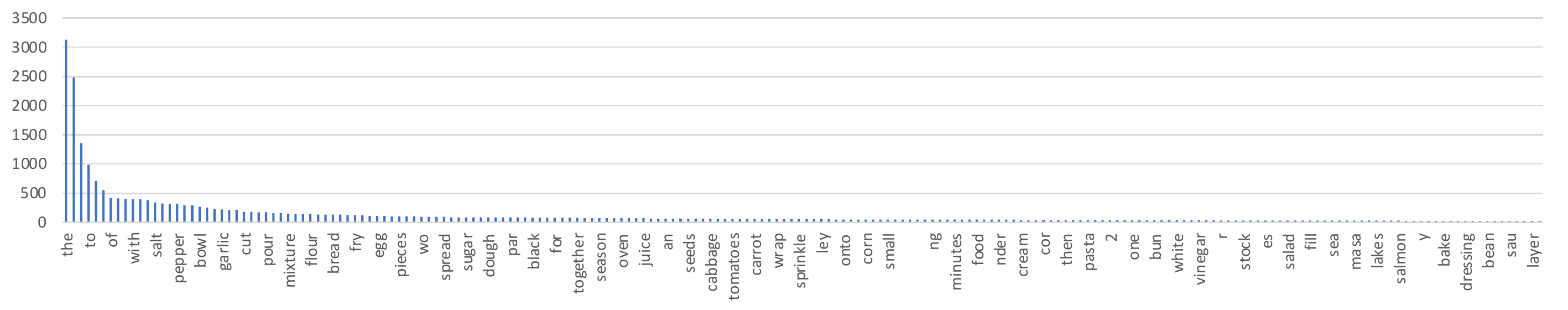}
\end{center}
\caption{Word frequency for the YouCook II evaluation set.}
\label{fig:youcook2wfreq}
\end{figure*}

\subsection{Results}
\label{results}

\subsubsection{Sentence Generation Automatic Evaluation}
\label{vid_cap_numbers}
For a thorough quantitative evaluation of the models, we utilize 8 established automatic evaluation metrics, BLEU (1-4) \cite{papineni2002bleu}, METEOR \cite{banerjee2005meteor}, ROUGE \cite{lin2004automatic}, CIDEr \cite{vedantam2015cider}, and SPICE \cite{anderson2016spice}. Results on YouCook II and ActivityNet are presented in \autoref{tab:cap_numbers}. For both datasets results using our implementation of the Transformer (T5) \cite{raffel2020exploring} for video captioning are also reported which is used as our baseline. Our proposed model, R3-Transformer consistently outperforms the baseline Transformer on both datasets. We also observe similar trend when comparing R3-Transformer with state-of-the-art methods. Please note that since we do not perform any language-vision pretraining, we only compare our model to train-from-scratch settings. Overall, these results provide evidence that the proposed model's performance improvement is generalizable across different domains.

\subsubsection{Analysis of Learned Relative Roles}
\label{role_analysis}
To further analyze the nature of information captured by learned \textit{relative roles}, which we refer as \textit{role}s for simplicity from now on, we conducted an experiment where the following probability is computed across all samples of the evaluation set:
\begin{equation}
\label{analysis1}
    \underset{\mathbf{k}}{max} \; p(\mbox{role \textbf{k} was selected} | \mbox{word \textbf{w} was generated}) 
\end{equation}
It represents for each given word how many times \textit{the same role} has been selected. Results on the full evaluation set of YouCook II are presented in \autoref{fig:youcook2roles}. 

These results suggest that most of the time the learned \textit{roles} are attracted to actions and predicates. This is consistent with our initial intuition for modelling where we leveraged the \textit{relationships} among roles and also among visual tokens (voxels). To make sure this phenomenon is not happening because of the actions/predicates having higher frequency in evaluation set, we performed a frequency analysis that is presented in \autoref{fig:youcook2wfreq}

Frequency analysis results in \autoref{fig:youcook2wfreq} shows that the observation of the learned \textit{roles} attracted to predicates is not due to the fact that they are more frequent words in the evaluation set and most probably is due to the specific way the proposed architecture is leveraging the semantic structure of the training data. 

\subsubsection{Further analysis based on predicates}
\begin{figure}[!ht]
\begin{center}
    \includegraphics[width=\linewidth]{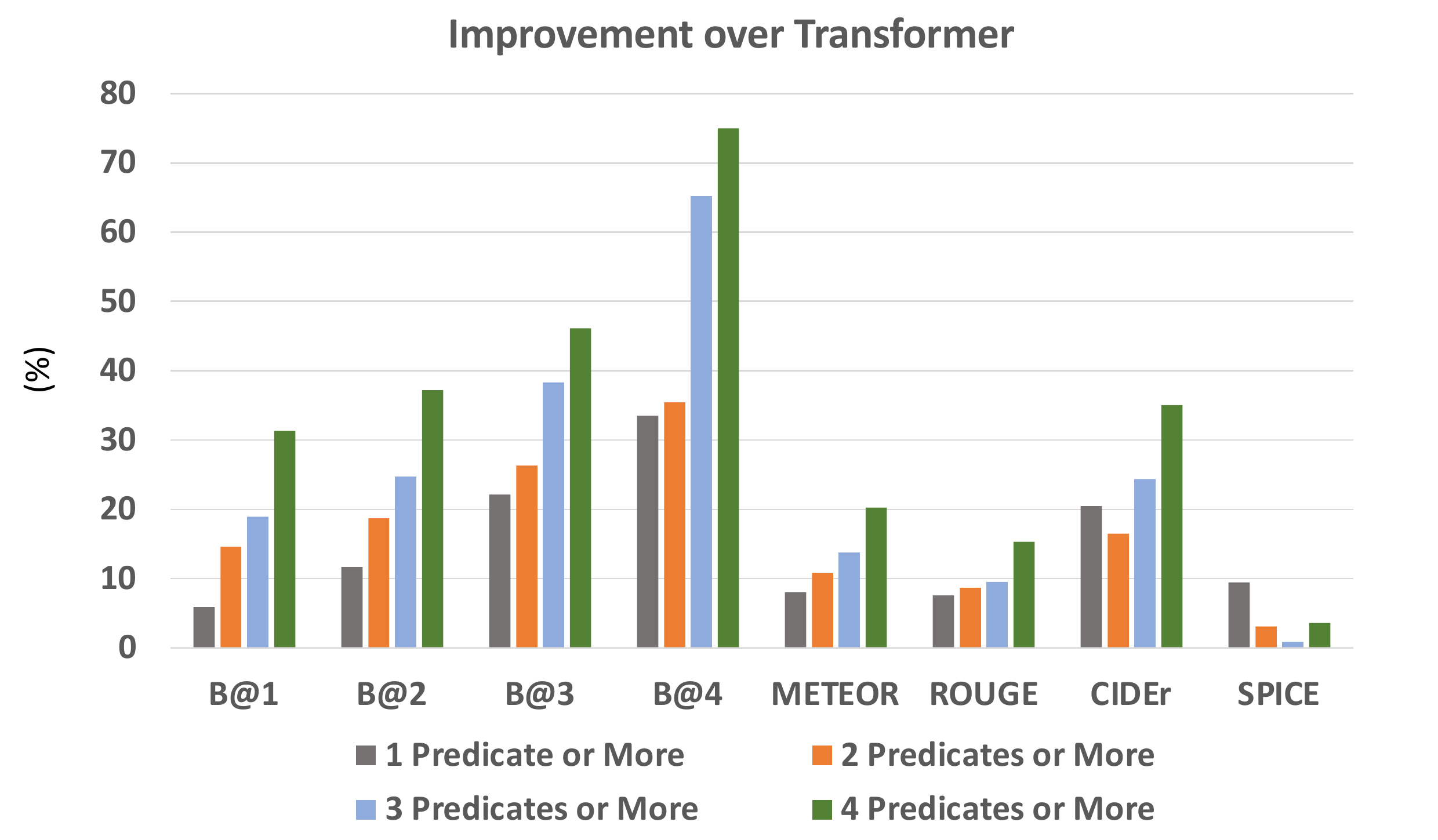}
\end{center}
\caption{Effectiveness of the proposed model in generating more challenging predicate-heavy captions on 4 different evaluation sets. Vertical axis represents the percentage improvement over baseline of the proposed model for different evaluation sets.}
\label{fig:predicate_heavy}
\end{figure}

The main intuition behind the \textit{special relativity} module is inducing a bias in the model towards the relative role that each token has with respect to its surrounding context. As we saw in section \ref{role_analysis}, these roles turn to be mostly visual actions, which serves as important information in generating a caption. In this section, we investigate whether \textit{predicates}, which also are among the learned roles, significantly contribute to the proposed model's performance. For this experiment we use a standard tagger \cite{tagger} to find Part Of Speech (POS) tags for all the ground truth captions in the YouCook II dataset. This will produce a POS tag per word in each caption, e.g., if the given word is a noun, verb, determiner, etc. We then merge different verb categories in the output of tagger to one category and refer to all of them as . We count the number of predicates per ground truth caption and sort the captions based on the number of predicates found by POS tagger. Based on predicate counts, we construct four subsets of evaluation set where we have at least 1, 2, 3, and 4 predicates in each set. The evaluation sets with a greater number of predicates are more challenging as they are usually longer captions with a more complex nature. Automatic evaluation is performed on each of the four evaluation sets separately. For each of the four sets, we then measure the normalized improvement percentage over the conventional Transformer. \autoref{fig:predicate_heavy} shows the improvement for each set, and as expected, the number of predicates in captions has a strong correlation with the amount of improvement that we gain over the baseline Transformer. This further verifies that R3-Transformer performs better in utilizing the role \textit{predicate} as a strong inductive bias in learning how to caption a video. This argument becomes more evident when we look at the word-level accuracy (BLEU metric) improvement, which are significantly larger compared to other metrics.

\subsection{Human Evaluation}
\vspace{-0.1cm}
We conducted human study using Amazon Mechanical Turk to compare the captions generated by R3-Transformer with the ones by Transformer. 200 video segments were randomly sampled from each of the tests sets of YooCookII and ActivityNet datasets. For each video segment, the human annotators were presented with the video url and the captions generated by R3-Transformer and Transformer.\footnote{We shuffled the two captions each time to avoid any order bias.} Each video segment was annotated by three annotators. We asked the annotators to compare the two captions on the following criteria:
\begin{itemize}
\small
    \item which caption is more relevant to the video
    \vspace{-0.2cm}\item which caption captures the right actions/objects in the video
    \vspace{-0.2cm}\item which caption is a fluent English sentence
\end{itemize}
Given two captions A and B, we asked the annotators to choose from the following options:
\begin{itemize}
\small
    \item Caption A is much better than caption B
    \vspace{-0.3cm}\item Caption A is slightly better than caption B
    \vspace{-0.3cm}\item Caption B is slightly better than caption A
    \vspace{-0.3cm}\item Caption B is much better than caption A
\end{itemize}

\begin{figure}[t]
\begin{center}
\includegraphics[width=7cm]{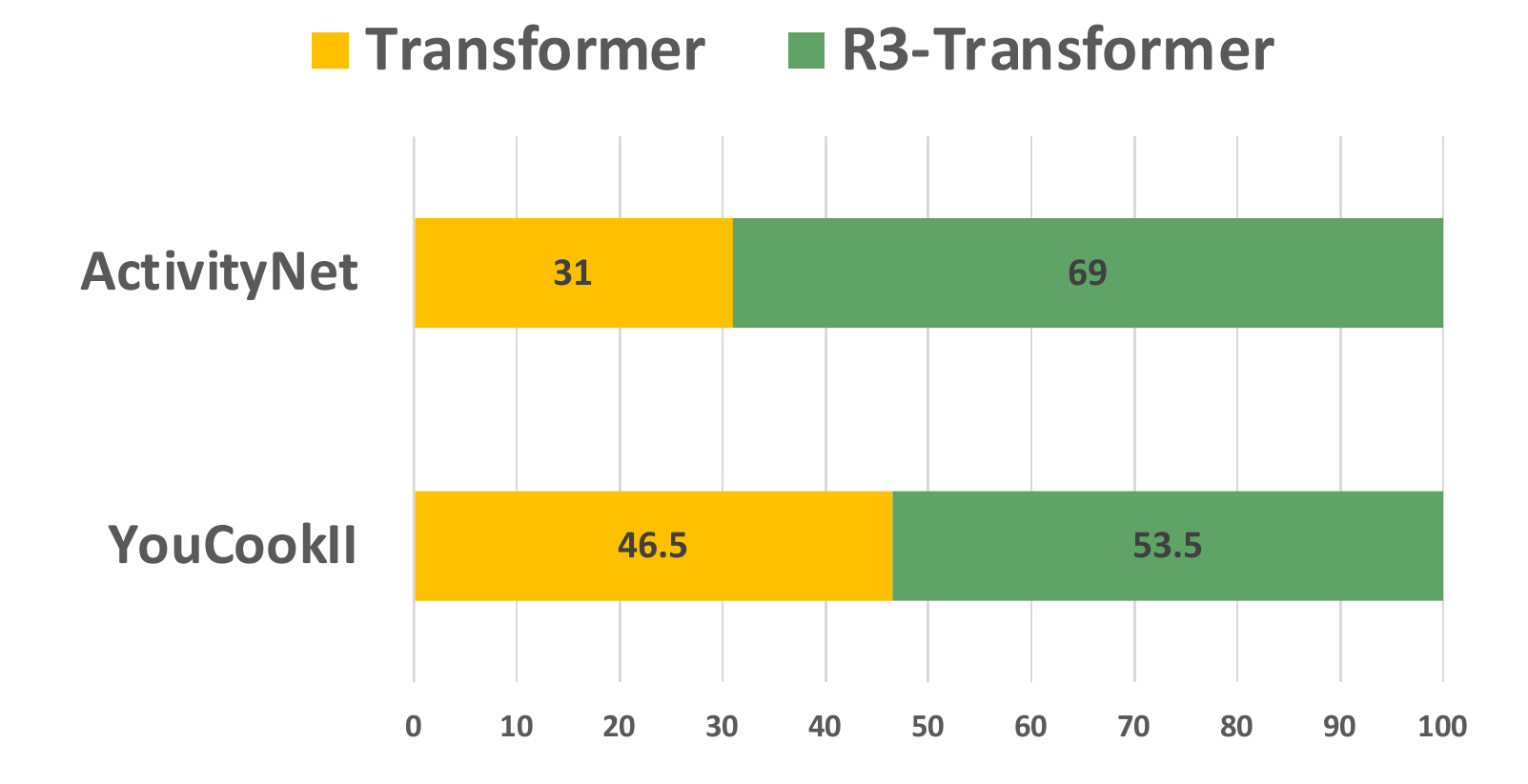}
\caption{ Results of pairwise comparison, as judged by human evaluators, between captions generated by Transformer and R3-Transformer on 200 video segments each from the test set of YouCookII and ActivityNet datasets. }\vspace{-0.5cm}
\end{center}
\label{fig:human-eval-results}
\end{figure}
\autoref{fig:human-eval-results} shows the results of this comparative analysis. For the video segments in ActivityNet dataset, human annotators preferred the captions generated by Transformer 31\% of the time whereas they preferred the captions generated by R3-Transformer 69\% of the time. For video segments in the YouCookII dataset, human annotators preferred the captions generated by Transformer 46.5\% of the time whereas they preferred the captions generated by R3-Transformer 53.5\% of the time. For more details about human evaluation refer to \ref{appndx:human_eval}.
\vspace{-0.1cm}
\section{Conclusion}
\vspace{-0.1cm}
\label{sec:conclusion}
In this work we proposed the use of a specific type of neuro-symbolic structure for video captioning. Our intuition was based on the assumption that for each given dataset there are limited number of predicates (\textit{roles}) that should be \textit{disentangled} from the rest of learned representations in the model. To achieve this we designed both a visual encoder and textual decoder to leverage the relationships among visual tokens (voxels), among words and among each other. We then used these relationships to define an attention mechanism over the \textit{roles} in our model's dictionary look up. Through experiments on two video captioning datasets from different domains, we made the case for better performance of the proposed model. Human evaluations were also conducted on both datasets and the results verified the improvement in generated captions' quality. Our preliminary analysis on learned \textit{roles} showed that most of the time they are attracted to predicates and action words.

\vspace{-0.2cm}
\subsubsection*{Acknowledgments}
\vspace{-0.2cm}
We would like to thank Nebojsa Jojic for his constructive discussions during the course of the project.

\pagebreak

{\small
\bibliographystyle{ieee_fullname}
\bibliography{egbib}
}

\newpage
\appendix
\section{Appendix}
Please refer to supplemntary materials.
\subsection{Details of human-based evaluations study}
\label{appndx:human_eval}

 
\subsection{Ablation Study}
\label{appndx:ablation}

\subsection{Captioning Examples}
\label{appndx:captions}

\end{document}